# Topic Modeling Genre: An Exploration of French Classical and Enlightenment Drama

Christof Schöch <christof_dot_schoech_at_uni-wuerzburg_dot_de>, University of Würzburg, Germany

## Abstract

The concept of literary genre is a highly complex one: not only are different genres frequently defined on several, but not necessarily the same levels of description, but consideration of genres as cognitive, social, or scholarly constructs with a rich history further complicate the matter. This contribution focuses on thematic aspects of genre with a quantitative approach, namely Topic Modeling. Topic Modeling has proven to be useful to discover thematic patterns and trends in large collections of texts, with a view to class or browse them on the basis of their dominant themes. It has rarely if ever, however, been applied to collections of dramatic texts.

In this contribution, Topic Modeling is used to analyze a collection of French Drama of the Classical Age and the Enlightenment. The general aim of this contribution is to discover what semantic types of topics are found in this collection, whether different dramatic subgenres have distinctive dominant topics and plot-related topic patterns, and inversely, to what extent clustering methods based on topic scores per play produce groupings of texts which agree with more conventional genre distinctions. This contribution shows that interesting topic patterns can be detected which provide new insights into the thematic, subgenre-related structure of French drama as well as into the history of French drama of the Classical Age and the Enlightenment.

## Introduction

The concept of literary genre is considered highly complex for several reasons. First, genres can be defined and described on a number of levels of description, such as plot, theme, personnel, structure, and style (for an introduction, see [Schöch 2017]), with style in turn concerning a range of aspects, such as spelling, lexicon, morphology, semantics, and syntax as well as rhythm, imagery, or complexity (for an overview, see [Herrmann et al. 2015]). Second, related genres are frequently defined on several different levels, such as length, form or theme: for example, subgenres of narrative prose fiction include the short story, the epistolary novel, and the libertine novel, creating overlap if not contradictions. Finally, consideration of genres as cognitive, social, and of course scholarly constructs with a rich history further complicate the matter, raising the question of what part formal features, on the one hand, and tradition and construction, on the other, play in the perception of literary genres (see [Hempfer 1973], [Schaeffer 1989]). 1

The wider context of this contribution is the early career research group on Computational Literary Genre Stylistics (CLiGS). There, we consider it of interest to analyze genres (and especially subgenres of drama and the novel) on a wide range of levels of description, stylistic as well as structural and thematic, and ranging from the use of function words all the way to plot structures. A particular focus lies on the question of how these various levels correlate and interact, and how they evolve over time in a given generic subsystem. The present contribution is one brick in that building, laying the focus on thematic aspects of genre and using Topic Modeling. This technique has proven to be useful to discover thematic patterns and trends in large collections of texts. Here it is applied, as has rarely been done so far, to collections of dramatic texts.[1] 2

In this contribution, Topic Modeling is used to analyze a collection (further described below) of French Drama of the Classical Age and the Enlightenment. This is a well-researched domain in French literary studies (for an account of the genre's formal poetics, see, for example, [Scherer 2001], and for a thorough overview, [Mazouer 2010/2014]). The general aim of this contribution is to assess whether Topic Modeling can be a useful, quantitative complement to established, qualitative methods of analysis of French dramatic texts from this period. More specifically, the contribution would like to discover: (1) what kinds of semantic types of topics appear in this collection, knowing that 3

Topic Modeling applied to fictional texts usually yields less abstract topics than when applied to non-fictional texts; (2) whether different dramatic subgenres have distinctive dominant topics, and if yes, whether those topics are expected or surprising, specific or vague, and abstract or concrete; (3) whether dramatic subgenres have distinctive plot-related topic patterns, and if yes, whether these patterns are dependent on subgenre or not; (4) and finally, to what extent clustering and classification methods based on topic scores produce results which agree with conventional genre distinctions.

## Hypotheses

Based on what we know about the history of French drama on the one hand, and the basic principles of Topic Modeling on the other hand, we can formulate a number of hypotheses or questions concerning the relations between topics and dramatic genres and subgenres. 4

First of all, dramatic subgenres such as comedy or tragedy being in part defined on the basis of their themes, and Topic Modeling bringing to the fore the hidden thematic structure of text collections, it can be expected that Topic Modeling applied to a collection of dramatic texts from a small range of different subgenres (comedies, tragedies and tragicomedies, in the present case) should bring out relatively strong genre-related topic patterns in the data. 5

More specifically, and because there is only a small number of subgenres in the collection, there should be a relatively large proportion of topics which are clearly distinctive of one of the subgenres involved. It will be interesting to note which of the topics will be most distinctive of comedies and tragedies: for instance, will they be clearly thematic topics, or will the semantic commonalities of topic words be of some other type? Will they be topics which we would expect to be characteristic of tragedies and comedies written in the seventeenth and eighteenth centuries (such as royalty vs. bourgoisie, honor vs. love, etc.), or will they be unexpected? It will also be of interest to investigate the topic-wise position of tragicomedy, which may either turn out to mix topics of both comedy and tragedy in a specific manner, or may also contain distinctive topics of its own. 6

Another aspect of the relations between topic and genre concerns plot. We know that on a very fundamental level, comedies and tragedies from the period studied here are distinguished by their typical plot patterns: tragedies tend to have a final act in which a lot of the protagonists are defeated or die, with conflictual power-relations and violent crime dominating. Comedies tend to have a final act leading up to one or several marriages, that is with a triumph of socially-accepted love relationships and happiness. If, as can be expected, there are topics related to such themes or motives, we should see a pattern across textual progression showing an increased importance of such topics towards the end of tragedies and comedies, respectively. 7

Finally, it is possible to invert the perspective from *a priori* categories and their distinctive characteristics to a data-driven, entirely unsupervised grouping of texts into (potentially genre-related) clusters. If topics turn out to be strongly distinctive of genre, then it can also be expected that clustering based on scores of topic proportions per play should result in groupings strongly related to genre. However, it remains to be seen whether such groupings confirm traditional genre-related divisions or diverge from them, and how they compare to groupings based not on topic scores, but directly on word frequencies. 8

## Data

The data used in this study comes from the *Théâtre classique* collection maintained by Paul Fièvre ([Fievre 2007-2015]). At the time of writing, this continually-growing, freely available collection of French dramatic texts contained 890 plays published between 1610 and 1810, thus covering the Classical Age and the Enlightenment. The majority of the texts are based on early editions made available as digital facsimiles by the French national library (BnF, Bibliothèque nationale de France). The quality of the transcriptions is relatively good without always reaching the consistent quality expected of more formally edited scholarly editions. In addition, the plays contain detailed structural markup applied in accordance with the *Guidelines* of the Text Encoding Initiative (TEI P4; for a general introduction, see [Burnard 2014]). For example, the plays' structure with respect to act and scene divisions as well as speeches, speaker names, stage directions and many other phenomena are all carefully encoded. In addition, detailed metadata has been added to the texts relating, for instance, to their historical genre label (e.g. *comédie héroique*, *tragédie*, or *opéra-ballet*) as well as the type of thematic and regional inspiration (e.g. French history, Roman mythology, or Spanish mores). A large part of this information can fruitfully be used when applying Topic Modeling to this text collection. 9

The collection of plays used for the research presented here is a subset of the *Théâtre classique* collection, defined 10

by the following criteria: Only plays from the period between 1630 and 1789 were included, because the collection contains only a comparatively small number of plays for the decades before and after these dates. Plays have only been included if they have between three and five acts, effectively excluding a certain number of one-act plays, in order for the plays to have a similar length and a comparable plot structure. Finally, all the plays included belong to one of the following subgenres: comedy, tragedy or tragicomedy. For most of the other relevant genres, only a relatively small number of examples is available. These criteria resulted in a collection of 391 plays (150 comedies, 189 tragedies, and 52 tragicomedies in verse or in prose), corresponding to approximately 5.6 million word tokens or 31 MB of plain text. For the purposes of this study, the plays have been split into smaller segments, resulting in 5840 documents used for Topic Modeling (see details below).[2]

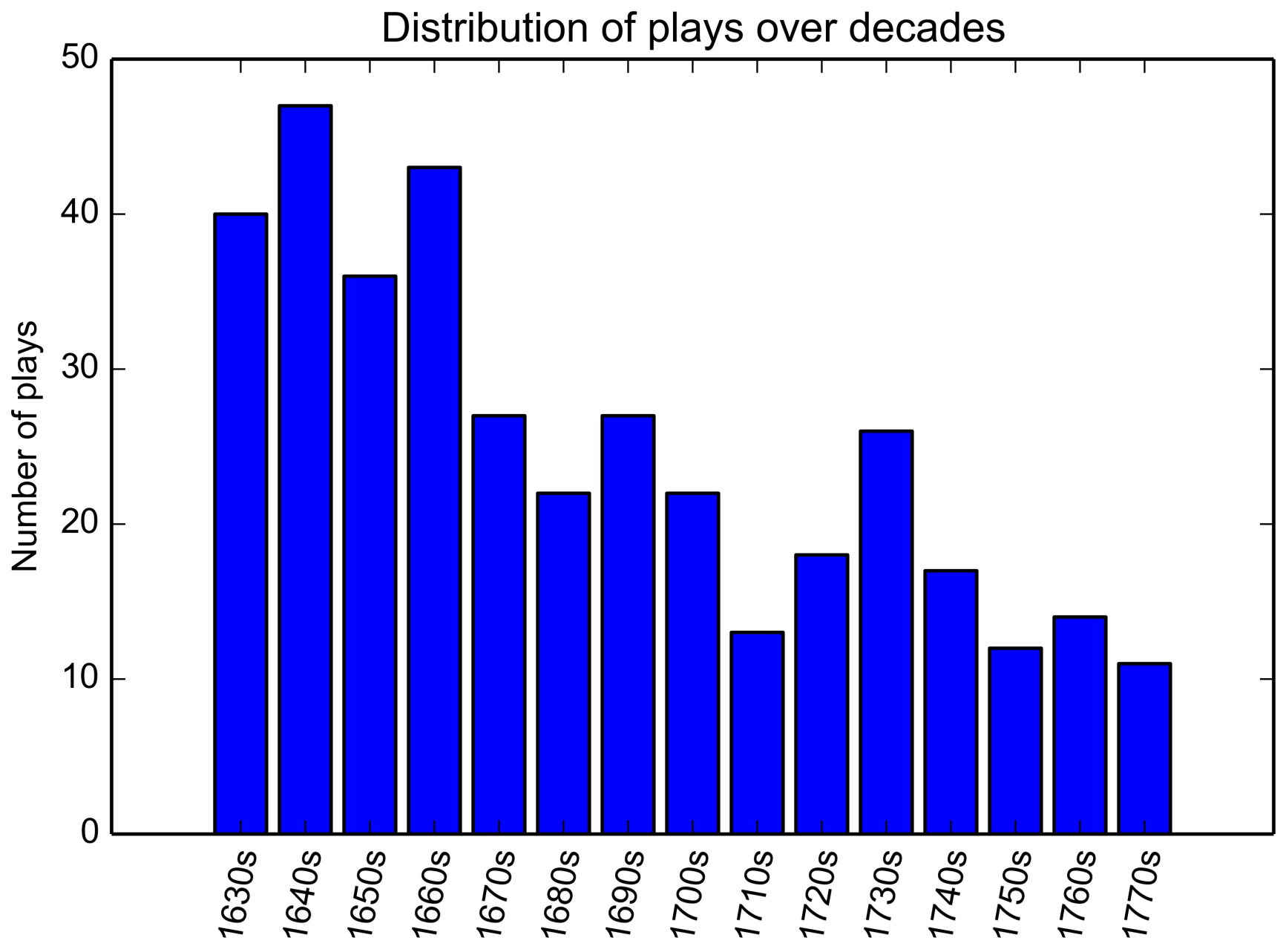

**Figure 1.** Distribution of plays in the collection used, by decades.

The plays' distribution over the time period covered by this study is shown in Figure 1. As can be seen, the coverage is not entirely balanced, with more plays from the seventeenth than from the eighteenth century, but there is a minimum number of 10 plays per decade. This collection corresponds neither to a random nor to a representative and/or balanced sample of all dramatic works produced during the period in question, which are more varied in length and more diverse in genre, and whose total number can only be estimated. To put the size of the present collection in perspective, one may consider the registry of known plays from the period 1620 to 1720 maintained by the *Théâtre classique* project, which currently lists 1914 plays for a 100-year period. 11

# Method

In the following, two aspects of the method used here are described. First, a brief explanation of Topic Modeling will introduce readers unfamiliar with the procedure to some of its most basic assumptions. However, Topic Modeling itself is just one step in a larger workflow involving both preprocessing and post-processing. Therefore, the more general processing workflow is also briefly described, with a focus on some of the decisions that need to be made with respect to several parameters used in the procedure. Also, a strategy for setting the parameters given a specific research goal is presented. 12

## Topic Modeling

Topic Modeling is an unsupervised method used to discover latent semantic structure in large collections of texts (for an introduction, see [Blei 2012]). In practice, individual words with the highest scores in a given topic are assumed to be semantically related words. This does not mean they must all belong to a common abstract theme (such as justice or biology). Especially in literary texts, it is also common for the shared semantic basis of words in a topic to be a particular setting (such as interiors or natural landscape), a narrative motive (such as horse-riding or reading and writing letters), or a social group of characters (such as noblemen or family members). However, the basis of similarity can also be some other aspect, such as the fact that all words are character names, or that all words come 13 from a certain register (such as colloquial words) or from a distinct language (such as Latin terms in otherwise French text). This fact somewhat qualifies the general assumption that the topics with the highest scores in a given text represent that text's major themes. Also, literary texts do not necessarily treat their dominant themes explicitly: unlike an essay or a research paper, a novel can be about social injustice, or a poem about death, without using these specific terms explicitly, showing them through concrete examples rather than explaining them through conceptual discussion.

On a slightly more technical level, a topic is a probability distribution over word frequencies; in turn, each text is characterized by a probability distribution over topics. Topic modeling is an entirely unsupervised method which discovers the latent semantic structure of a text collection without using lexical or semantic resources such as electronic dictionaries. This means that Topic Modeling is not only language-independent, but also independent of external resources with potential built-in biases. Rather, Topic Modeling is based on assumptions about language first developed in distributional semantics, whose basic tenet is that the meaning of a word depends on the words in whose context it appears. As John R. Firth famously put it in 1957, “a word is characterized by the company it keeps”. In line with this idea, the highest-ranked words in a topic are those words which frequently occur together in a collection of documents. A second, related assumption of Topic Modeling is a specific view of how the writing process is envisioned. In this view, text is generated from several groups of semantically related terms which are chosen, in different proportions for each text, when the text is written. Topic modeling reconstructs, based on the resulting text alone, which words must have been in which group and which probability they had of being selected in the writing process (see [Steyvers and Griffiths 2006]). Because this is a model with a very high number of unknown variables, it cannot be solved deterministically. Rather, it is solved with an initial random or arbitrary distribution of values which is then iteratively improved until a certain level of convergence between the texts predicted by the model and the actual texts in the collection analyzed is reached. This also means that the results of Topic Modeling a given collection, even with identical parameters, may not yield exactly identical results every time the technique is applied. However, experience with multiple repeated modeling processes show that generally speaking, and given enough iterations, it is a rather robust technique, with results varying in the details of word ranks rather than in the general topics obtained (see below). 14

The most commonly used implementation of Topic Modeling uses an algorithm called Latent Dirichlet Allocation ([Blei et al. 2003]), but there are several precursors (such as Non-Negative Matrix Factorization) and an increasing number of alternative algorithms. Also, besides the most commonly used tool, MALLET ([McCallum 2002]), which is written in Java, several other tools are available, such as gensim ([Rehuřek and Sojka 2010]) and lda ([Riddell 2014b]), both written in Python. Several extensions to Topic Modeling have been proposed, such as hierarchical Topic Modeling ([Blei et al. 2004]) and supervised or labeled Topic Modeling ([Ramage et al. 2009]), with both variants also being available in MALLET. Due to the availability of relevant tools and tutorials (e.g. [Graham et al. 2012] or [Riddell 2014a]), and because it answers to the wish of many scholars in the humanities to gain a semantic access to large amounts of texts, Topic Modeling has proven immensely popular in Digital Humanities (for applications, see e.g. [Blevins 2010] , [Rhody 2012] , or [Jockers 2013]) for analyses of different literary genres. 15

## The Topic Modeling Workflow

Topic Modeling has been performed as part of a larger processing workflow described in this section (see Figure 2 for an overview of the process). The workflow is almost entirely automated using a custom-built set of Python scripts called tmw, for Topic Modeling Workflow.[3] Starting from the original XML/TEI-encoded texts from *Théâtre classique* and making use of the structural markup, speaker text and stage directions have been extracted from the plays in order to exclude interference from prefaces, editorial notes, or speaker names. In the next step, as the plays are relatively few but have a considerable length, each play has been split into several smaller segments of similar length (around 1000 words). For the research reported here, arbitrary segment boundaries have been preferred to segmentation along the act and scene divisions, as this would result in overly heterogeneous segment lengths. This results in an average of 14.9 text segments of comparable length per play, or 5840 segments in total.[4] 16

Lemmatization and morpho-syntactic tagging have been performed using TreeTagger ([Schmid 1994]) with a language model and tagset specifically developed for historical French in the framework of the PRESTO project ([PRESTO 2014]). [5] Lemmatization, i.e. the transformation of each word form to its base form such as could be found in a dictionary entry's head word, has been performed because French is a highly inflected language, and differences in the word use in different inflectional forms may obscure the common semantic structure which is of interest here. Morpho-syntactic tagging identifies the grammatical category of each word token and not only allows to filter out (speaker) names mentioned by other speakers (which are not of interest here), but also allows for the 17 specific selection, for Topic Modeling, of only a certain number of word categories. For the research presented here, only nouns, verbs and adjectives have been retained for analysis, under the assumption that those are the main content-bearing words, while all other word categories, which mostly contain function words, have been excluded from the data.

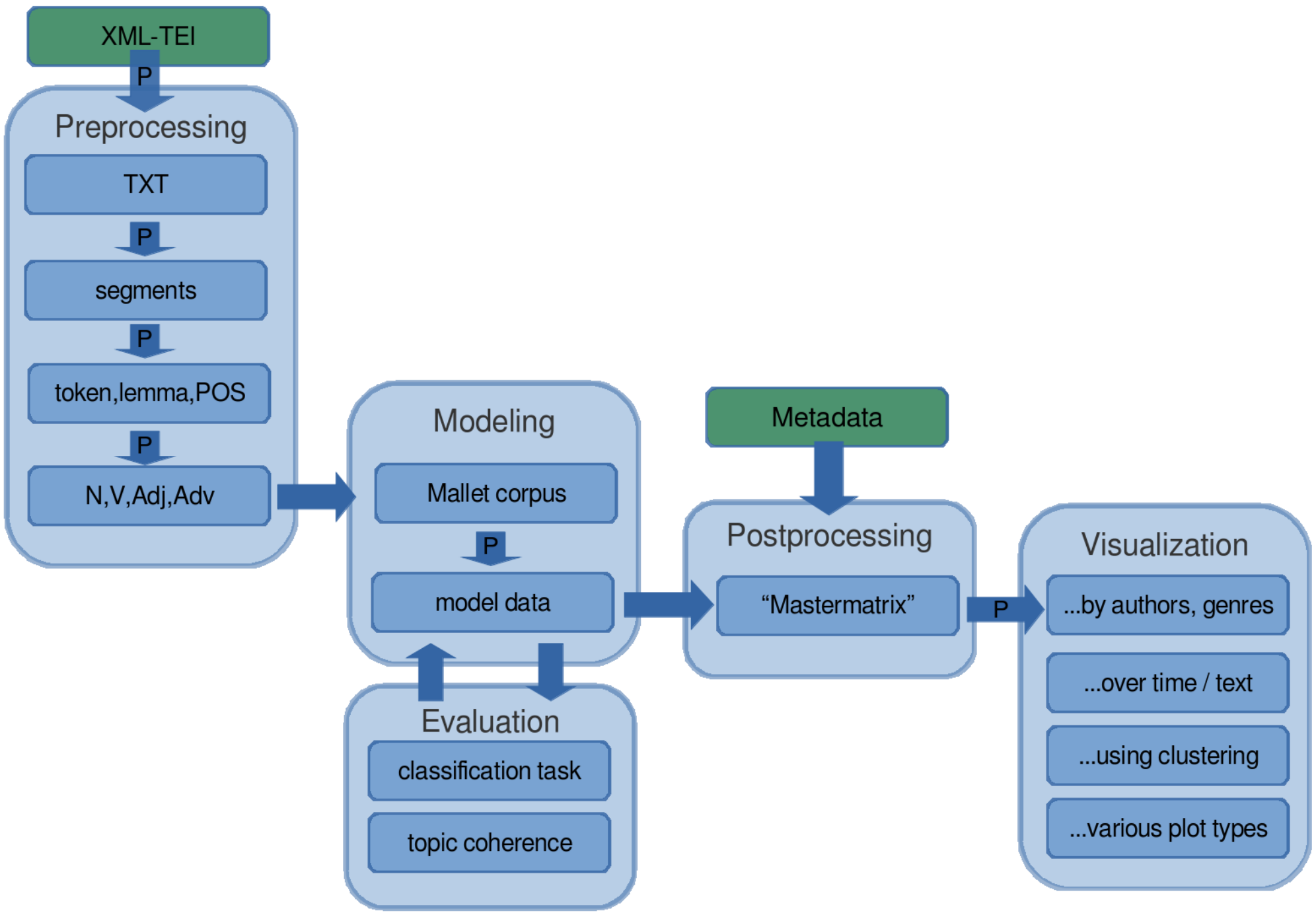

**Figure 2.** The Topic Modeling Workflow (tmw) as implemented in Python.

Ultimately, this means that instead of 391 XML-encoded entire plays, 5840 short pseudo-text segments made up of sequences of noun, verb, adjective and adverb lemmata have been submitted to the Topic Modeling procedure. Each of these segments can be identified as to the individual play it belongs to (which, in turn, is associated with descriptive metadata, such as the date of publication, the author, and the subgenre of the play) and as to the relative position in the play it comes from (with a granularity of five subsequent sections of the plays, corresponding roughly, if not structurally, to the five acts of the majority of the plays included). 18

The Topic Modeling procedure itself has been performed using MALLET. After (in this case) 6000 iterations, the result of this process is a topic model of the initial data, represented, among other things, by a table of all topics, ranked according to their overall probability in the entire text collection, with a ranked list of words most strongly associated with each of them; by a table containing the probability score of each topic in each of the 5840 text segments; and by a table containing the probability score of each word in each topic. 19

There are many parameters influencing the result of the modeling process, but the following are of particular importance: The number of topics (what level of granularity should the model have?), the hyperparameters alpha and beta (affecting the distributional profile of the words in each topic and that of the topics in each document, respectively; see [Wallach et al. 2009] and [Buntine and Mishra 2014] for details), the number of iterations (how many times should the model be updated?), and the optimization interval (at which intervals, in terms of iterations, should the hyperparameters be adjusted automatically? see [Schöch 2016]). For all of these parameters, there doesn't seem to be any hard-and-fast rules on how to determine an appropriate setting, other than a firm understanding of the topic modeling procedure combined with a good knowledge of the dataset under scrutiny and a clear research objective which the model is supposed to support. In order to overcome this difficulty of largely intuitive and arbitrary decisions, a series of 48 different models have been created based on a range of parameter settings, systematically varying the number of topics (six levels: 50, 60, 70, 80, 90, 100) and the hyperparameter optimization interval (eight levels: 50, 100, 300, 500, 1000, 2000, 3000, None) while keeping the number of iterations constant (at 6000 iterations). In order to evaluate the model and decide which model should be used in the remainder of the study, a machine-learning task has been defined which consists of classifying the plays according to their (known) subgenre. This task appears to be appropriate since the main focus of the present study lies in the thematic differentiation of the dramatic subgenres present in the collection.[6] The data relating to the probabilities of 20 each topic in each play, for each combination of parameters, has been used as input to this task and the performance evaluated in a ten-fold cross-validation setting using four different machine learning algorithms (Support Vector Machines, k-Nearest Neighbors, Stochastic Gradient Descent and Decision Tree). The performance of each algorithm for different input data is shown in Figure 3 . As can be seen, the subgenre classification task is solved with an accuracy of around 0.70-0.87 (the baseline being 0.48). Although differences between the best-performing models are slight, the best results are obtained by the SVM for the model built with 60 topics and the optimization interval set at 300 iterations, with a mean accuracy of 0.87. Therefore, this is the model which has been used in the remainder of this study.[7]

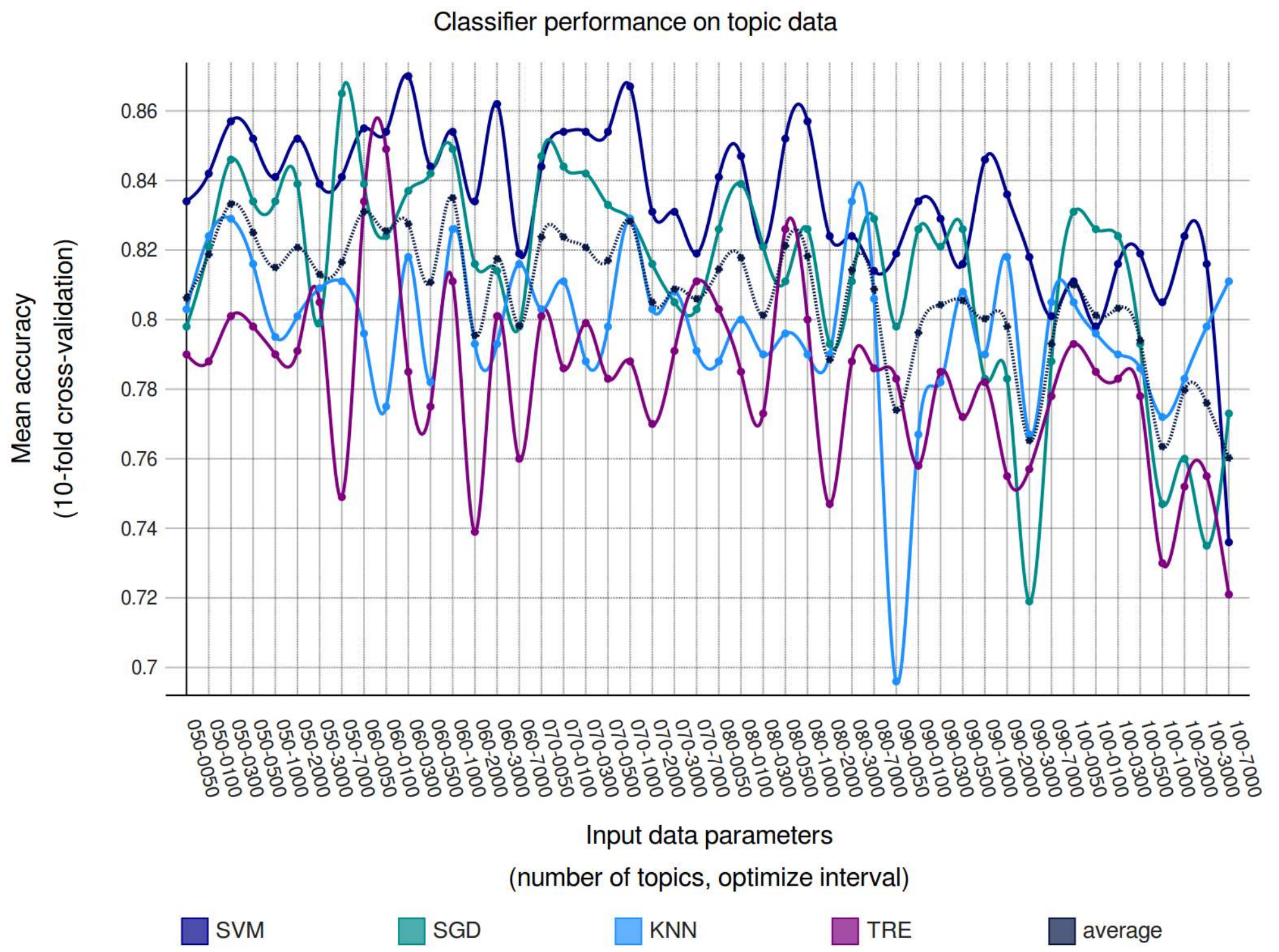

**Figure 3.** Results from a classification task using topic probability data as input for subgenre classification and four different classifers.

It should be noted that even with such precautions taken, some degree of intuitive or arbitrary decisions remains in the procedure and may have more or less direct influence on the results obtained. For this reason, it is all the more important to document the choices made. With this in mind and in order to increase the transparency of this research as well as allow the results to be reproduced, the dataset as it has been used, the Python code employed in the workflow, the descriptive data pertaining to the 48 different models and a set of graphs for the model selected for further analysis, have been published in the CLiGS projects repository on GitHub.[8] 21

## Results and Discussion

In the following three sections, three types of results will be discussed and related to the hypotheses described above. First of all, results relating to the topics found in the collection of plays. Then, results pertaining to topics which are distinctive for the three subgenres contained in the collection, including results relating to genre-specific plot-related patterns. Finally, results from clustering and classification based on topic scores as well as raw word frequencies will be presented. 22

### Topics: Structure and semantic coherence

An initial inspection of the 60 topics obtained with their top-40 words, visualized as word clouds, shows that most of the topics display a relatively high level of (subjective) coherence. A first selection of topic word clouds is shown in Figure 4. One common effect of a highly asymmetric topic probability distribution is that the topics with the highest probability scores (i.e. the ones most widely represented in the collection) are less interpretable than most others, in the sense that they are rather generic and vague. However, the model used here is notable for the absence of such topics; even the most widely present topics are relatively well-defined: Here, topic 32 is the highest-ranked topic; it has an overall score of 0.244 and its top five words are “*coeur*, *amour*, *aimer*, *oser*, *madame*” (heart, love, to love, to 23 dare, madam), describing a perspective on the omnipresent love theme marked by love as a challenge and a reward. Another highly-ranked topic is topic 3; it has an overall score of 0.172 and its top five words are “*secret*, *connaître*, *craindre*, *cacher*, *apprendre*” (secret, to know, to fear, to hide, to learn), resulting in a topic concerned with epistemological issues. Topics with very low probability scores (i.e. those appearing only in a few plays) are typically highly specific, but tell us less about the collection as a whole. Again, this phenomenon is comparatively less prevalent in the model used here, but it can still be noted for topics 30 and 15, which are both of a very low rank. Topic 30 has a score of 0.024 and its top five words are “*vers*, *auteur*, *pièce*, *beau*, *esprit*” (verse, author, drama, beautiful, spirit/wit) and is a metafictional topic centered on producing aesthetic literary works. Topic 34 has an even lower score of 0.014 and its top five words are “*mal*, *monsieur*, *médecin, guérir*, *malade*” (hurting, mister, doctor, to recover/heal, sick). Both topics are precisely focused and interesting, but occur only in very few plays or a single author. For instance, the sickness-topic just mentioned (topic 34) is strongly present only in two comedies, in *L'amour médecin* by Molière and in *Élomire hypocondre, ou les médecins vengés* by Le Boulanger de Chalussay, and a lot less strongly in a small handful of other plays. If one happens to be interested in one of these very specific topics, Topic Modeling provides a great way of identifying plays which should be included in a more detailed analysis. The most relevant topics for the research presented here, however, are those with less extreme probability scores, because subgenre distinctions are located by definition somewhere between individual plays and an entire collection of plays. Using just 60 different topics and a relatively low optimization interval provides a maximum of such topics of mid-range importance in the collection.

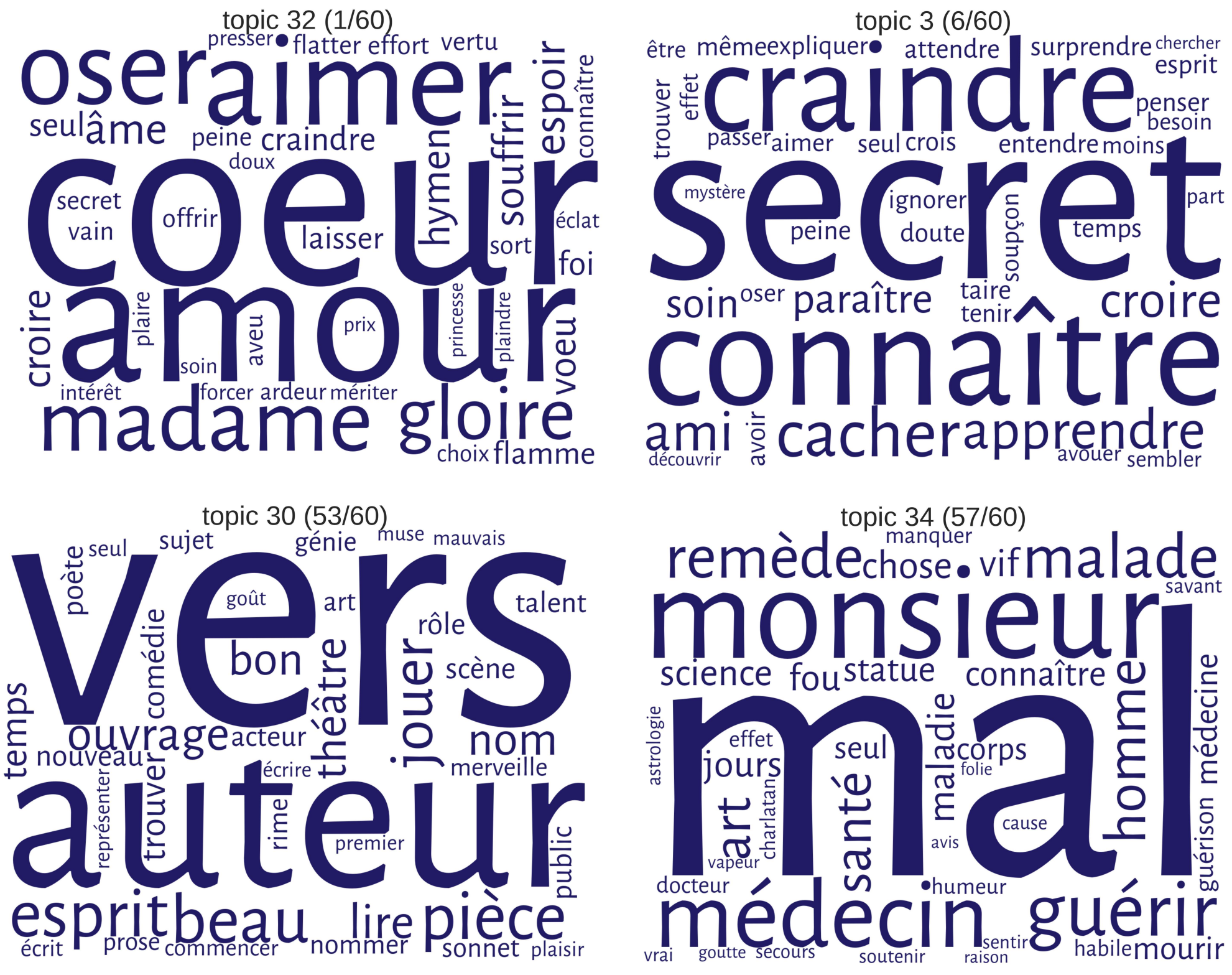

**Figure 4.** Word cloud visualization for selected topics: high and low overall topic probability (larger font size indicates higher word probability; the numbers in brackets indicate the topic rank).

The selection of topics in Figure 5 shows another phenomenon, related to the internal structure of topics (which in 24
part depends on the alpha hyperparameter). Most topics show a small number of quite important words (i.e. with high probability in the topic, displayed in a very large font size), with a relatively smooth drop-off and a long tail of less important words (displayed in very small font size). However, some topics show a different internal structure: for instance, in topic 48, only “*père*” and “*enfant*” (father, child) has a very large score, with subsequent words much less important in that topic. The effect is even more marked in topic 27, in which only a single word, “*ciel*” (sky), has a very high score, with an extremely clear drop in scores for all other words in the topic. Inversely, topic 1 has a large number of words with relatively high scores, the first five being “*monsieur*, *argent*, *bon*, *payer*, *écus*” (mister, money, good, to pay, écus). The same phenomenon can be observed in topic 50. The word cloud visualizations nicely bring out this internal structure of the topics.

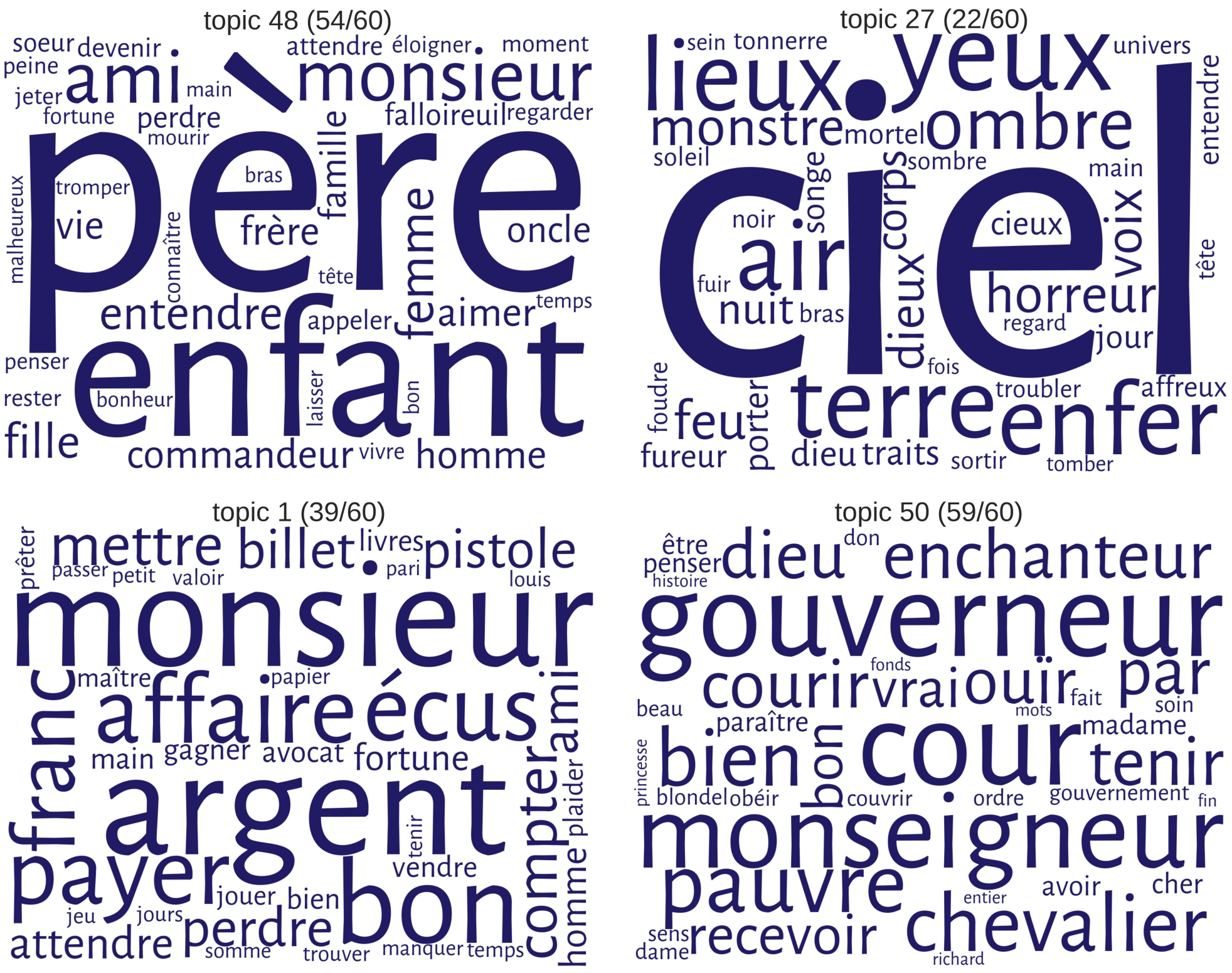

**Figure 5.** Word cloud visualization for selected topics: sharp and soft decline in word probabilities.

What, then, are topics characteristic for this collection of plays, what are the themes most commonly found in them? 25
Figure 6 shows examples for several types of topics. Many of the topics found are related to clear, abstract themes, such as love, death, crime and marriage, which are also themes we can expect to appear in plays of the seventeenth and eighteenth centuries. One such topic is shown here for illustration, namely topic 18, clearly related to life and death: “*mourir*, *mort*, *douleur*, *vivre*, *vie*” (to die, death/dead, pain, to live, life). Such topics typically come from the upper region of the probability scores. Others focus on the dramatic personnel, such as topic 5 related to family members: “*fils*, *père*, *mère*, *frère*, *enfant*” (son, father, mother, daughter, child). Quite a number of topics are rather more concrete, such as those related to a setting, as in topic 46: “*beau*, *berger*, *bois*, *arbre*, *lieu*” (beautiful, shepherd, wood, tree, place). Others focus on quite specific activities, such as topic 30 mentioned above or topic 26 devoted to reading and writing letters: “*lettre*, *lire*, *écrire*, *billet*, *main*” (letter, to read, to write, note, hand). These topics typically come from a somewhat lower range of probability scores. The latter types of topics, related to activities, setting and personnel, show that taking a method such as Topic Modeling, developed initially for collections of non-fictional prose such as scholarly journal articles or newspapers, and adapting it to the domain of literary texts, actually changes the meaning of the word “topic”: to put it another way, the “topics” found in literary texts are not only abstract themes such as justice, knowledge or crime, but also more concrete activities typically performed by fictional characters, like writing, eating, drinking, hunting, speaking and thinking, or literary settings or motives such as the sea, the forest, or interiors. While the presence of the former is related to the choice of including verbs into the analysis, the same is not true for the latter. Note that what becomes visible here are different *types* of semantic coherence, irrespective of the *degree* of semantic coherence as assessed by various topic coherence measures (for an overview, see [Röder et al. 2015].

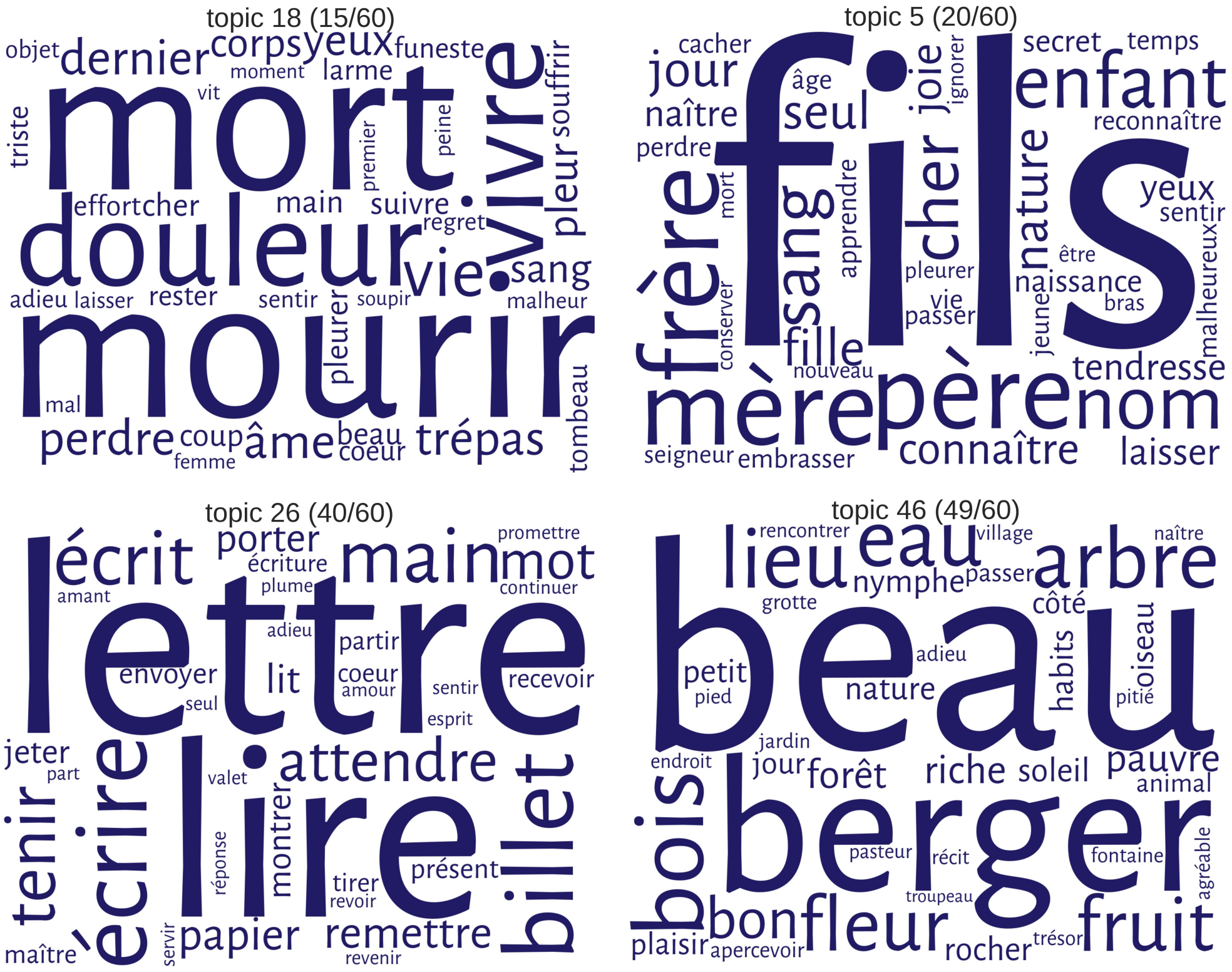

Figure 6. Word cloud visualization for selected topics: abstract and concrete types of topics.

In a small number of cases, several topics are related to very similar semantic fields, despite the fact that with just 26
60 topics, the granularity of the model is already comparatively low. This indicates that there may be a hierarchical structure to the topics, and that it may be useful to think of topics in a hierarchical relationship or as clusters.[9] This is the case for topics concerning family members and relations as well as for death-related topics (not shown), and especially for no less than 8 topics related in some way to love, and containing among their top-ranked words items such as “*amour*” and “*coeur*” (love, heart). Four such topics are shown in Figure 7. Interestingly, the words following these top-ranked words differ significantly in each of these topics: for instance, in topic 6, words like “*ingrat*, *trahir*, *perfide*, *venger*, *crime*” (unthankful, to betray, insidious, to avenge, crime) link this kind of love to ideas of intensely negative emotions. Conversely, topic 24 prominently contains words such as “*aimer*, *âme*, *beau*, *yeux*, *doux*” (to love, soul, beautiful, eyes, soft), linking this kind of love to pleasing outer appearance as well as inward beauty. What seem to be very similar topics, at first, turn out to contextualize the top words in very different ways. These differences may turn out to be related, in addition, to subgenres, because some of these several love topics are associated more strongly with one subgenre than with others, an issue that will be addressed in the next section.

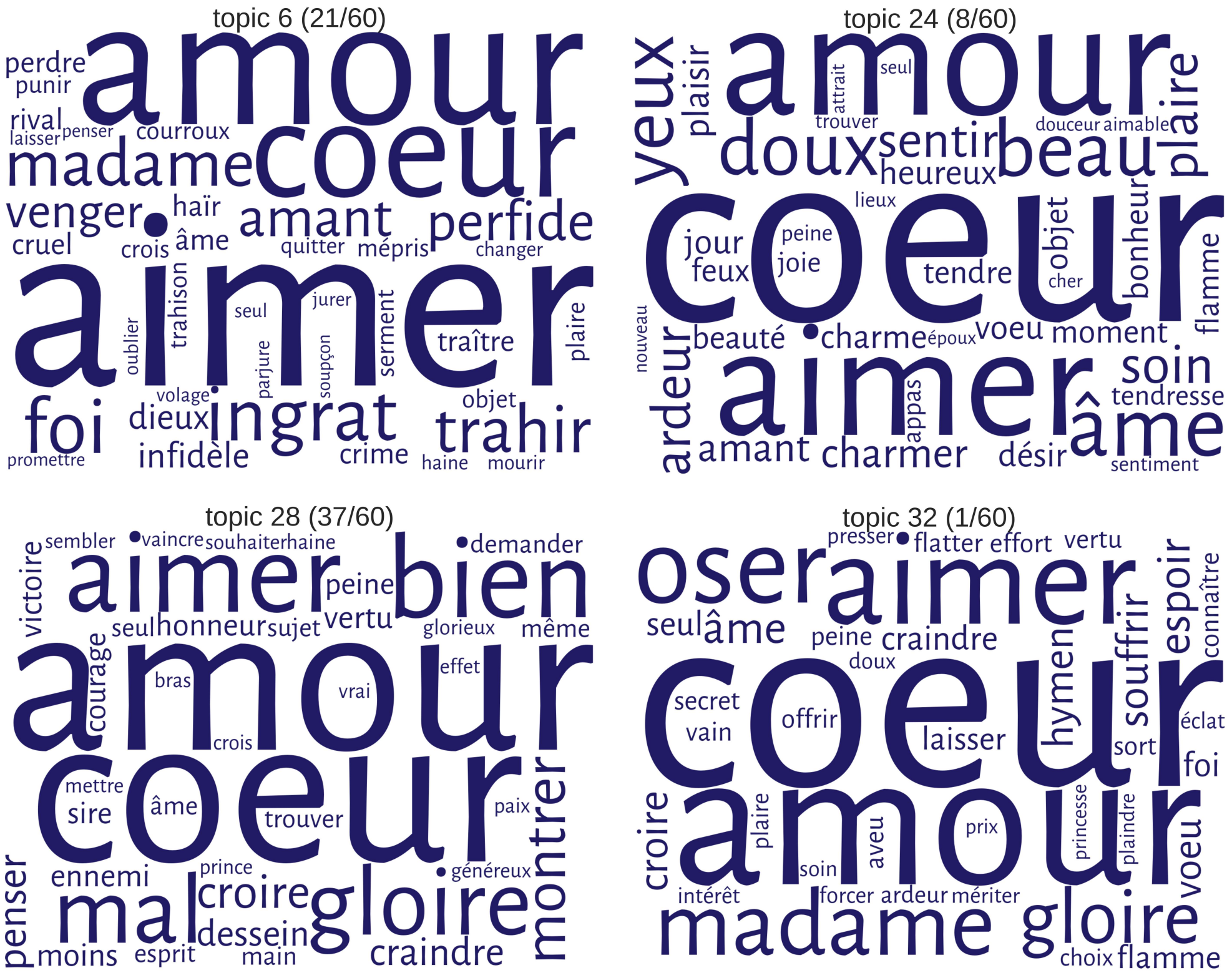

**Figure 7.** Word cloud visualization for selected topics: topics related to love.

## Topics and genre: distinctiveness and plot-related patterns

While it is possible (and interesting) to extract, from the data, information showing which topics are over-represented 27
or under-represented in certain authors or in certain decades, this paper focuses on the relation between topic and dramatic subgenres. One way to discover such distinctive topics is to proceed as follows. The topic scores obtained for each text segment are aggregated and their mean (or median) is calculated, taking into account the genre of the play that each text segment belongs to. This yields, for each topic, a score for its average importance, technically speaking its mean (or median) probability in each of the three subgenres. The topics with the highest probability in a given subgenre are not necessarily also the most distinctive ones, i.e. those which are over-represented in one subgenre relative to other subgenres, because some topics are just highly present in all subgenres. Therefore, a topic-wise mean normalization is performed, effectively creating positive values for topics over-represented in a given subgenre and negative values for topics under-represented in a given subgenre. In order to identify the topics with the most extreme differences between topics, the topics are then sorted by decreasing topic-wise standard deviation across the three subgenres. In this way, the topics with the highest variability across subgenres can be displayed, which are also the topics which are the most distinctive of different subgenres.[10] Figure 8 shows a heatmap of such subgenre-related average topic scores for the twenty topics with the highest cross-genre variability.

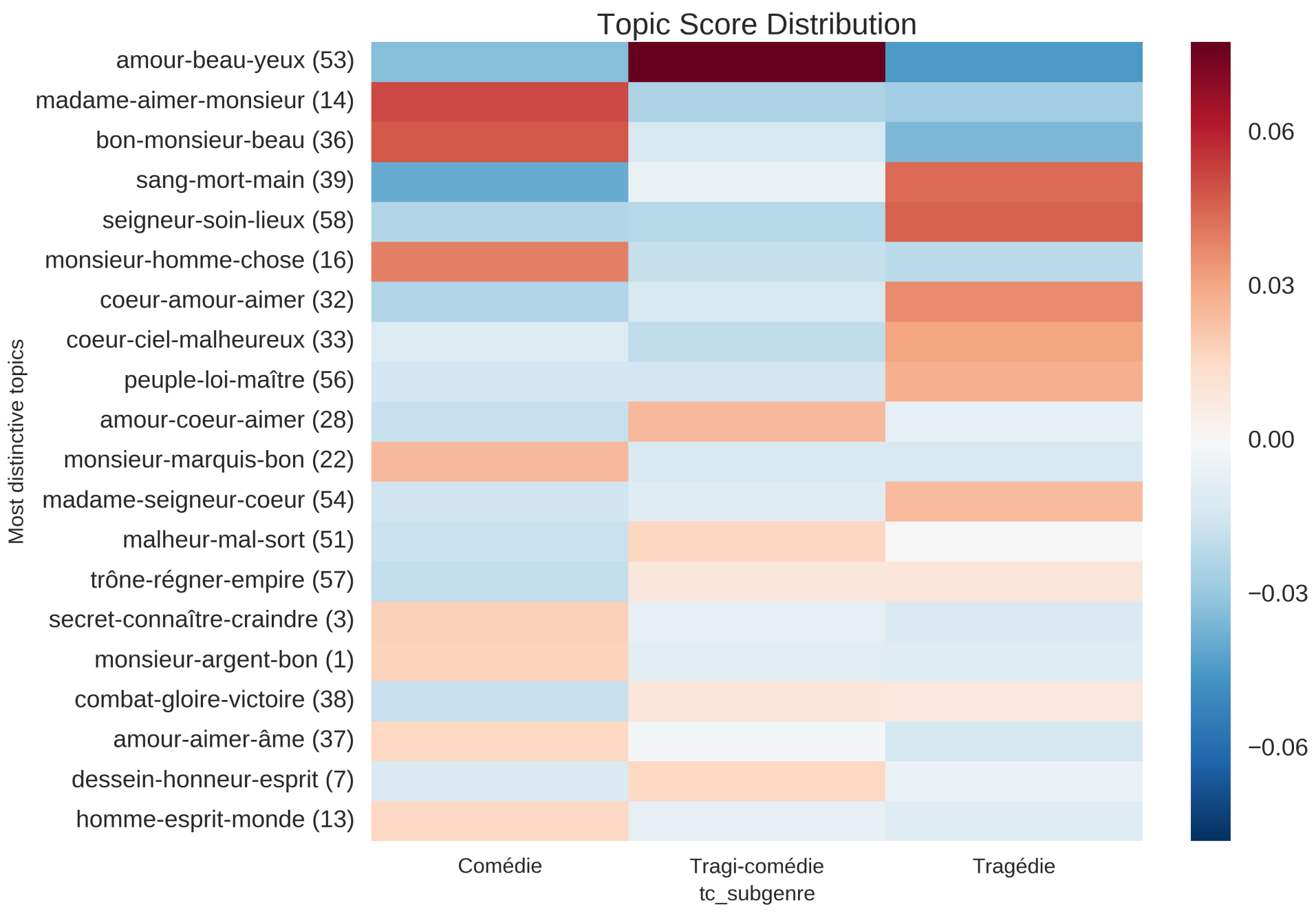

**Figure 8.** Heatmap of top-20 topics by subgenre, sorted by decreasing standard deviation across subgenres (red indicates over-representation, blue indicates under-representation).

As can be seen, each subgenre has several distinctive topics, i.e. topics which in one genre have a score significantly higher both in relation to the other genres for the same topic (across rows), and to other topics for the same subgenre (across columns). The two most distinctive topics for comedies, with their top-ranked words, are topic 14 ("*madame*, *aimer*, *monsieur*" / madam, to love, mister) and topic 36 ("*bon*, *monsieur*, *beau*" / good, mister, beautiful), relating somewhat vaguely to social interactions. The most distinctive topics for tragedy are topic 39 ("*sang*, *mort*, *main*" / blood, death, hand) and topic 58 ("*seigneur*, *soin*, *lieux*" / sire, care, place). The first of these topics is much more immediately thematic than the two top comedy topics and relates unequivocally to violent, physical, and deadly crimes, something which seems to indicate that the tragedies analyzed here fulfill, to a significant extent, rather stereotypical genre expectations. The second is a bit less interpretable, but appears to be related to courtly negotiations and intrigue. Tragicomedy, interestingly, only has one very clearly distinctive topic, topic 53 ("*amour*, *beau*, *yeux*" / love, beautiful, eye) and one a bit less so, topic 28 ("*amour*, *coeur*, *aimer*" / love, heart, to love). This seems to indicate that at this level of analysis, tragicomedy is a mix of both tragedy and comedy rather than a genre of its own, something which confirms established knowledge about the genre, but does not yet give us more detailed information about which of the two genres is related more closely, topically and hence, thematically, to tragicomedy. Does tragicomedy have more topic-based overlap with comedies or with tragedies? (The next section will attempt to give an answer to this question.) Finally, all three subgenres each have at least one quite distinctive topic related to "love", i.e. topics in which either "*amour*", "*aimer*" or "*coeur*" are among the top two words. As has already been seen, each of the "love" topics actually represents quite a different perspective on the theme of love, when looking at some of the words in the top twenty range or so (see Figure 7 , above). 28

To summarize, not only are very different topics associated with different subgenres, but they also seem to be different types of topics: rather vague topics concerning personal relations and interactions and love for comedy, and quite focused, abstract topics related to crime and politics for tragedy. This seems to indicate that while tragedy remains defined by the topics it addresses, possibly quite explicitly, for example in monologues, comedy does not have such clear, abstract defining topics, although marriage, money and deception (topics 23, 14 and 3) do seem to be somewhat distinctive. Other expected themes like humor for comedies or vengeance for tragedies do not appear prominently among the topics. Also, some topics which seem quite similar at first glance but are distinctive, one for comedy, one for tragedy, actually reveal deep-running differences which explain their association with different subgenres. 29

While topic-related patterns across subgenres, then, are very clearly present, the question remains whether this is also the case for plot-related patterns. This question can be investigated in the following manner: for each topic, topic scores are aggregated not by grouping the values of each topic in each of the 5,840 text segments only by their genre association, but also by the section of the play they belong to. This was done here with a relatively rough 30 granularity of five sections per play corresponding, numerically if not structurally, to the five acts of most of the plays, and for comedy and tragedy only. In addition, instead of calculating mean values per genre and section, all individual scores are taken into account so that the boxplots display key aspects of the distribution of the topic scores. In this way, the rising or falling importance of a given topic over the course of the comedies or tragedies can be assessed and compared with existing knowledge about characteristic features of plot in these subgenres.

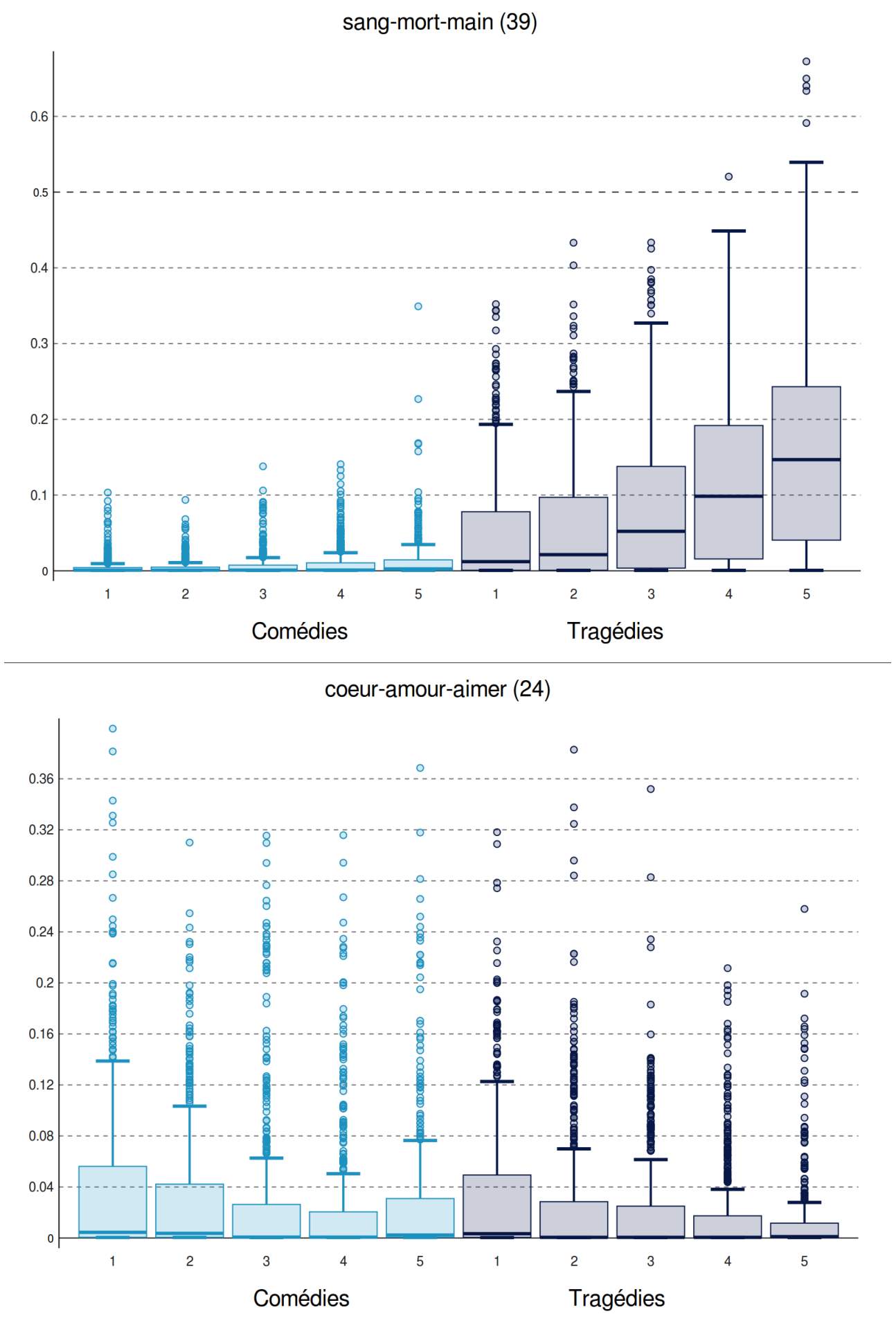

**Figure 9.** Distribution of topic scores over the course of average tragedies and comedies: topics 39 and 24.

Overall, while many topics do not show strong changes in importance over the course of plays in one or several 31
subgenres, approximately one fifth of the topics in the model analyzed here do. In several cases, a topic rises or falls in one genre while remaining stable in the other. This is the case, for example, for topic 39 (another death-related topic) and obviously related to tragedy (see Figure 9). Not only is this topic indeed much more prevalent in tragedy than in comedy, but it also gains importance over the course of many tragedies: in the first two fifths of the tragedies, its median is close to zero, with a considerable amount of outliers, i.e. single text segments with an unusually high score. (This is an effect of the many segments in which a given topic plays only a negligible role, given that each segment analyzed here is very short and only accommodates a few of the sixty topics). However, in the remaining three fifths of the tragedies, the median gradually raises to 0.05 (in section 3), 0.98 (in section 4) and 0.15 (in section 5; the underlying distributions differ very significantly). Quite obviously, the fact that tragedies frequently end with threats of cruelty and actual death manifests itself in the topic scores here. A similar pattern, but only in comedies, can be found for topic 0 related to marriage and indicative of a happy ending (not shown). Similar raising patterns can be also be found for topic 18 and 21 for either tragedy or comedy (not shown). A falling pattern can be seen for topic 24, a love-related topic, and while the decrease over the course of the plays is less marked than in the previous examples, the same pattern appears to be present in both comedies and tragedies (only some underlying distributions differ significantly). Because there are several other love-topics, this does not necessarily mean that love is of decreasing importance over the course of the plays; rather, different types of love-related topics appear to be of importance at different times in many of the plays. Other topics with patterns of falling importance over the course of the plays are topics 13, 22, 56, 57 and 58 (not shown).

A number of plot-related trends exist in the data, then, some stronger than others, and some of which lend themselves more easily to interpretation than others. When these kinds of patterns concern only one genre but not the others, they may easily be overlooked when looking only at all plays together. Therefore, this approach is relevant on a methodological level, because this is an example of a genre-related characteristic which only becomes clearly visible when looking at the development of topic importance over the course of many plays separated into subgenre-related groups. 32

## Topic-based clustering

The differences observed so far rely on the *a priori* subgenre classification of plays, accepting the historical subgenre labels as given. This, it may be argued, is problematic not only because the labels and their use may not be uncontroversial, but also because some seemingly significant differences in topic score distribution across such a small number of categories may always be found, no matter what the categories are. Moreover, if the aim is to find out how the dramatic genre as a whole is structured internally into subgroups of similar plays, such preconceived categories are not helpful. To explore whether the topics are indeed structuring the collection along the lines of subgenres, the perspective should also be inverted and a topic-based, unsupervised clustering method be performed on the data. This also allows one to move away from predefined, historical, potentially problematic genre labels, which allow discovering distinctive topics but may obscure other structure in the data, be it topic-based commonalities between several subgenres or additional divisions within one subgenre. 33

Several approaches to this clustering task can be taken, for example Principal Component Analysis (PCA) or Hierarchical Clustering (HC), among others. In any event, such a clustering step performs another data transformation on the topic scores. These topic scores, however, are already the result of a dimensionality reduction process, since LDA takes the very high-dimensional word vectors representing the documents and transforms them to the much lower dimensionality of the topics. (Note that both PCA and HC can also be applied directly to the word vectors instead of taking the intermediary step of LDA). In fact, PCA is in some ways quite similar to LDA in that it transforms the vectors to a new space of lower dimensions and may not perform optimally on data that has already undergone a similar procedure ([Joliffe 2002]). However, the advantage of PCA is that the topics most strongly correlating with a given principal component can be determined through the “loadings” of that component, something which aids the interpretation of the resulting model. Also, the position of documents to each other can be visualized in a two- or three-dimensional plot which allows for considerable nuance. HC builds a distance matrix out of the feature vectors representing the distance (or degree of dissimilarity) of every document to every other document. This and the following steps of creating a linkage matrix and visualizing it as a dendrogram is a somewhat distinct type of process from LDA (and PCA) and may for that reason be preferred when using data that has already undergone a dimensionality reduction. Also, clusters and subclusters of the data can be more readily inspected based on the distances at which more or less large clusters and subclusters are merged. Rather than choosing among these two analyses, both have been performed and their results are compared here. 34

Figure 10 shows the result of a PCA based on the topic scores. Here, the unit of analysis is the play, and the features characterizing each play are the mean probability scores of each of the 60 topics. The scatterplot displays the two components in the data which summarize the greatest amount of variation in the data.[11] 35

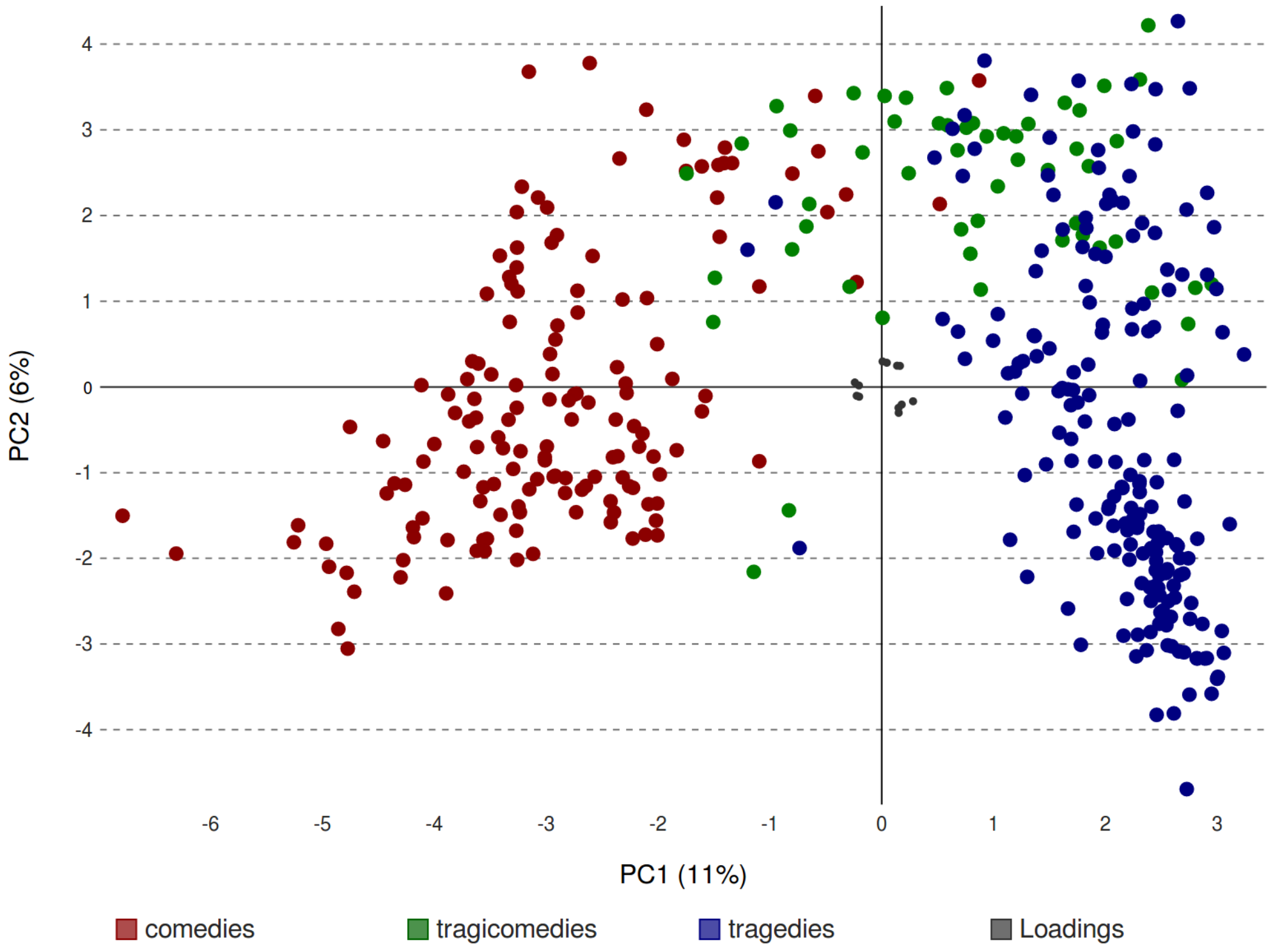

**Figure 10.** PCA plot of plays based on 60 topic scores. Tragedies are shown in blue, comedies in red and tragicomedies in green. (Click here to download an interactive plot.)

Each circle in the plot represents one play, and their relative proximity or distance indicates topic-based, thematic similarity or difference in the three dimensions shown. The colors of the circles correspond to the conventional genre labels of each play, which however do not influence the positions of the circles. The coloring only allows us to see to what degree the topic-based similarity of the plays corresponds with their conventional genre label. This correspondence is very high, and the first component (PC1, 11%, horizontal axis) clearly contributes the most to a separation of the plays into comedies (red circles to the left) and the tragedies (blue circles to the right). [12] It is true that to some extent, this may be an effect of correlations between authorship and genre (some authors, like Voltaire, having predominantly written only in one subgenre), but many authors in the collection have written plays in several subgenres. The second component (PC2, 6%, vertical axis) separates the tragicomedies from the rest of the plays, at least to some extent, with the tragicomedies being found only in the upper center region of the plot. However, there is substantial overlap between tragicomedy and tragedy as well as some degree of overlap between tragicomedy and comedy. When clustering plays based on their topic scores, then, tragicomedy appears to be more closely related to tragedy than to comedy. This fits one conventional description of tragicomedy as a tragedy that ends well, i.e. as a type of play with similar personnel, plot and themes as tragedy, except that there is no disastrous ending. Another interesting feature of the PCA plot is that there seem to be at least two distinct groups of tragedies. These groups cannot be simply explained by chronology, and the prose/verse distinction is also not pertinent for tragedy, for which virtually all examples are written in verse (an exception is discussed below). 36

Overall, we see that there are two major, well-defined groups of tragedies and comedies, but also secondary groups of tragedies as well as overlap between tragedies, tragicomedies and comedies. This supports the idea of a theory of genres as prototypes that can be realized in more or less complete ways (see e.g. [Hempfer 2014]). A thorough investigation into what unites the plays of each subgroup and what distinguishes them from the other subgroup, thematically and/or otherwise, will need to be performed in further research. 37

The second strategy intended to yield an insight into how the topics structure the 391 plays into distinct groups is Hierarchical Clustering. Cluster Analysis has been performed using the Euclidean distance metric and the "Ward" method of agglomerative clustering. The correlation coefficient achieved was 0.447. 38

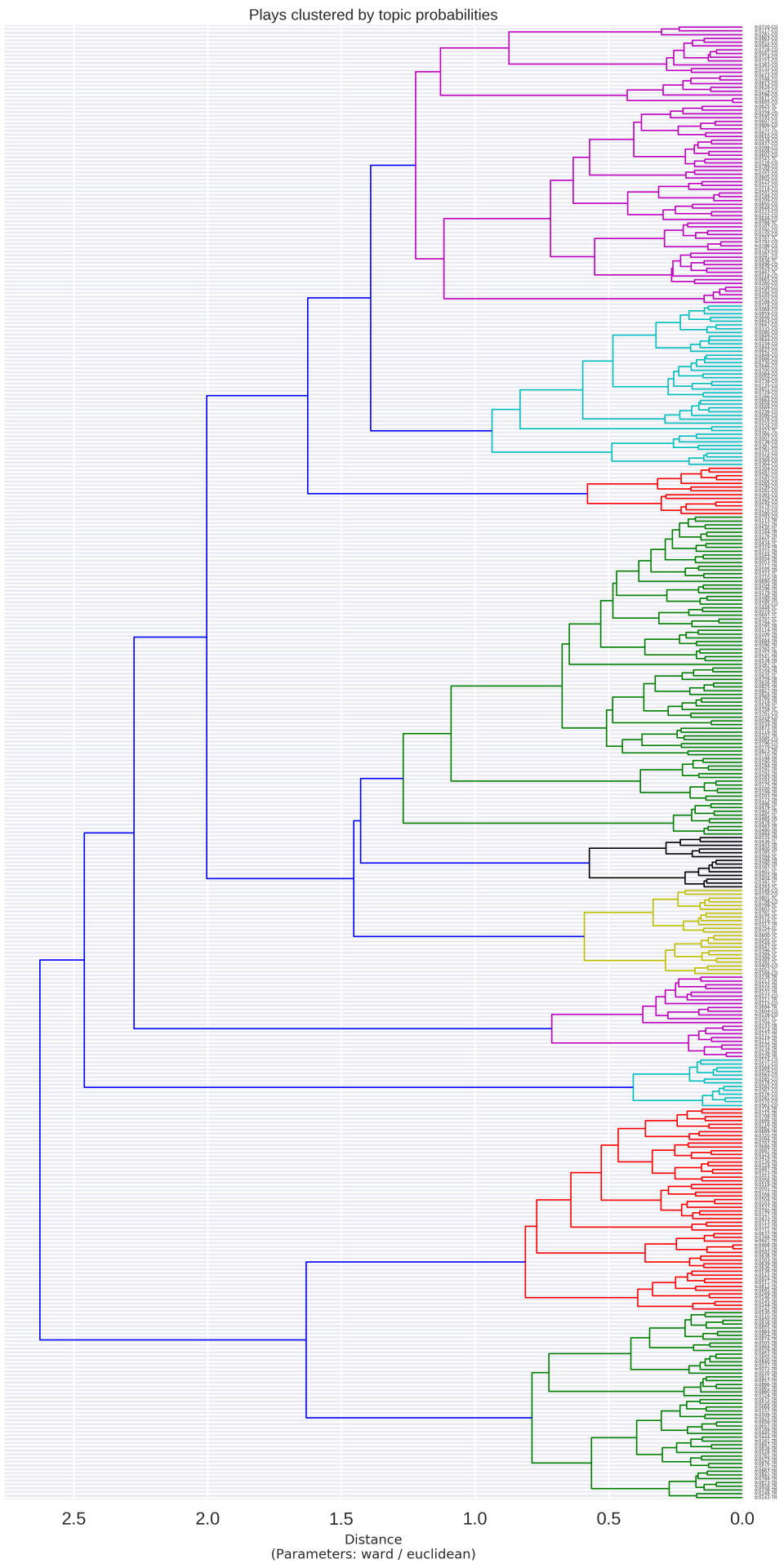

**Figure 11.** Hierarchical Clustering dendrogram of plays based on 60 topic scores (Ward method with Euclidean distance metric).

The cluster dendrogram (Figure 11) shows similar plays or clusters of plays being connected at low distance values. The labels to the right indicate the identifier of the play as well as its conventional genre category ("TR" for tragedy, "CO" for comedy, "TC" for tragicomedy). The colors of the clusters do not correspond to genres, but to clusters whose members are joined beyond the (arbitrarily set) threshold value of 1.3, resulting in 10 distinct subclusters. We can see that among these subclusters, the green and red ones at the bottom of the graph are populated exclusively by tragedies. The smaller turquoise cluster just above them is made up of comedies only, and they appear quite distinct from the remaining plays. The three magenta, turquoise and red subclusters at the top of the dendrogram correspond almost exclusively to comedies. In the middle of the dendrogram, there are several more mixed clusters, notably containing most of the tragicomedies along with other plays. The black subcluster in the middle of the dendrogram, for example, is an almost even mix of tragedies and tragicomedies, while the yellow subcluster just below it contains mostly comedies and tragicomedies. In a manner similar to the PCA-based scatterplot, the dendrogram shows that while some tragedies and comedies form quite large distinct groups, the tragicomedies mingle partly with the comedies, partly with the tragedies. 39

The two tragedy clusters at the bottom of the graph do not contain a single play of a different subgenre label.[13] But the occasional exception can be found in the three upper comedy clusters, where a small number of tragicomedies appear. One interesting case from among the more mixed clusters is *Socrate*, written by Voltaire in 1759 (identifier: tc0872) and a rare specimen of a tragedy in prose, something which may explain its position separate from most other tragedies. In addition, it appears closest to another play written in prose, an allegorical play in which the personified tragedies of Voltaire enter into a dialogue, something which appears to produce a play with topic proportions similar to the actual Voltaire play. In addition, although this play is labeled as a tragedy, it is certainly a very unusual tragedy and its historical subtitle is in fact “*comédie*”. 40

Overall, the results from clustering using PCA confirm that even when the algorithms do not know about subgenre categories, but cluster the plays only based on their topic scores, that is relating to their thematic similarity in a broad sense, subgenre turns out to be a prominent factor entering into the resulting clusters. 41

## Topics or words?

As we have seen, topic-based clustering, both in the form of Principal Component Analysis and Hierarchical Clustering, appears perfectly able to use the information about the topic probability scores to create groups of plays that are in line, to a very significant proportion, with conventional subgenre labeling of the plays in the *Théâtre classique* collection. In addition, we have seen at the beginning of this paper that topic-based classification into subgenres, when evaluated on those same conventional subgenre labels using ten-fold cross-validation, performs with a mean accuracy of between 70 and 87 percent, depending on data and classifier used. It is true that the tragedies, with 189 of the 391 plays, constitute a significant proportion of the plays in the collection. This means that the classification baseline should be considered to be no lower than 189/391, that is 48.3 percent, which corresponds to the proportion of true positives one would obtain when simply always choosing the category of the largest group. Still, these high levels of accuracy are a considerable feat, given the highly controversial nature of subgenre distinctions, even for a strongly conventionalized genre such as Classical and Enlightenment French drama. 42

However impressive these results may be, the question remains whether topic-based classification or clustering, because it is based on information highly relevant to genre distinctions, outperforms established alternative approaches, such as classification based on simple most-frequent words as commonly used in stylometry and particularly in authorship attribution (for an overview, see [Stamatatos 2009]). It turns out that when comparing results from classification based on topic scores on the one hand and word frequencies on the other, performance for both methods is on a similar level, with a slight advantage for the word-based approach. For both approaches, standard classifiers from the scipy package (k-Nearest-Neighbor, Decision Trees, Stochastic Gradient Descent and Support Vector Machines) have been used with very little fine-tuning. In the case of the topic probabilities, the model with 60 topics and a moderate optimization interval of 300 iterations yielded an accuracy of 87.0% when using SVM (with a linear kernel), an improvement of almost 40 percent points compared to the baseline (see Figure 3 above). However, when using word frequencies transformed to z-scores and using a word-vector of the 3500 most frequent words, the best-performing SVM achieved an accuracy of 90.5 percent (see Table 3 in the annex). This is not much above the topic-based classification, but the results are robustly above 90% even with word vectors of a length between 2500 and 4500, while the accuracy for the topic-based methods drops rather more quickly and unpredictably with slightly different parameters. 43

This result is surprising, because both methods rely on different views of the original text of the plays: while the topic scores have been obtained based only on content-bearing words and are abstractions from individual word tokens, the frequencies of the 3500 most frequent words used here have not been lemmatized and also contain a large number of function words. Such function words used to be associated with authorship rather than genre or subgenre, and could have been expected to skew results or represent noise with regard to genre. However, recent research has also shown that there is no simple separation of authorship and genre signals based on the word frequency spectrum (see e.g. [Rybicki and Eder 2011], [Kestemont et al. 2012] or [Schöch 2013] on the issue of authorship and genre in stylometry). In addition, the author signal is likely muted here due to the large number of authors in the dataset compared to the very small number of subgenres. 44

# Conclusions

To summarize the major results of this study, one may state first that, as far as the topics obtained are concerned, most topics obtained with the parameters chosen through the classification task are highly interpretable. Some of the topics are clearly thematic, others are related to the character inventory of plays, to recurring dramatic actions or to setting. Future work could usefully establish a more systematic approach to the types of topics, ideally leading to an empirically-founded typology of topics. Such a typology would allow to describe major genres such as narrative fiction, dramatic works and poetry by the prevalence of their types of topics (rather than specific individual topics), something which would allow better comparison across collections and even languages. 45

Also, based on metadata regarding the subgenre of each play in the collection, the strength and nature of topic-related patterns across subgenres can be observed, with each subgenre having a number of clearly distinctive topics. In addition, some topics show plot-related trends over the course of an average comedy or tragedy, trends which can in many cases be meaningfully related to existing knowledge about the subgenres' plot structure, especially the happy ending in comedy and the deadly finales of tragedy. 46

Clustering based on the topic scores obtained also yields results which can be usefully related to subgenres and show that the distinctiveness of topics with regards to genre is not a projection but an actual pattern in the data. 47

Overall, it appears that interesting patterns can be detected using topic modeling, patterns which, in many cases, confirm existing knowledge about the themes and plot of subgenres on the basis of a significantly larger amount of data than can be taken into account in close-reading approaches. In some cases, surprising results provide new insight into the history of French drama of the Classical Age and the Enlightenment or provoke new hypotheses which may need further inquiry in the future. This is the case, particularly, of the finding that there may be three relatively distinct groups of tragedies as well as two groups of comedies.

It has been found in previous (as yet unpublished) studies by the author of this paper that strong genre signals exist in the collection explored here on the levels of function words, content words and syntactic structure. It is interesting to see that they also exist on the level of theme or topic. Once each of these (and more) levels of description of genre have been analyzed in more detail for the collection of French plays, new insights into the structure of the subgenres and into their development over time will become possible by analyzing the ways in which these different levels of description correlate, interact, or possibly contradict each other. Ultimately, these results also show that far from merely being a projection or a social construct, literary genres do have a textual reality that can be assessed quantitatively. It is on the basis of such a reality and within the potentialities and constraints it offers, that different views of literary genre as a concept, and of literary history as the continual evolution of literary genres, become possible. 48

It would be tempting to finish this paper on such an optimistic note. However, it seems necessary to also point to some of the challenges connected to research of the kind presented here. First of all, many questions remain open and while some results can readily be linked to established knowledge in traditional literary scholarship, other phenomena remain to be explained. For example, the subgroups of tragedies and comedies found in the PCA scatterplot and Cluster Analysis dendrogram will need to be investigated in future work. Also, the results obtained for plot-related patterns are promising, but the inclusion of three-act plays as well as plays with a prologue, in the collection used here, may have obscured plot-related patterns in the majority of five-act plays. However, confirming established knowledge or widely accepted hypotheses on the basis of larger datasets is useful in itself, especially at a time when there is only a limited amount of experience with techniques newly applied to the domain of literary texts. This helps build confidence in results which may support entirely new hypotheses. A second, more technical, challenge is related to the methods used here to obtain values for topics depending on the subgenre of a play or the section in a play. It may be more appropriate, and may produce better results, to either perform logistical regression on the data to discover associations between topics and genres (or other categories), or to include the section information into the modeling process from the start, which would effectively mean practicing "sequential LDA" ([Du et al. 2006]) or "topics over time" ([Pölitz et al. 2015]). Tentative experiments with labeled LDA ([Ramage et al. 2009]; [Blei and McAuliffe 2008]) have been inconclusive so far. Also, as can be seen from the close relations between a number of topics, hierarchical LDA ([Blei et al. 2004]) may be able to capture existing structure among the topics themselves. A third, less technical but no less important challenge is the fact that it would still be advantageous to have larger number of texts available. 391 plays certainly is a substantial number, even compared to the total production of the time (see section “Data” above). Also, it is certainly more than a single researcher could read with benefit inside a limited time period. However, considering that the study covers a period of 150 years and three subgenres, 391 plays are actually not that many. For example, this effectively means there are only around 8 plays per decade and subgenre, on average, included in the collection (not considering the uneven distribution over the decades). Also, several significant subgenres had to be discarded from the analysis, because only a handful of examples of them are currently available in the *Théâtre classique* collection. This is a challenge that only continued efforts in high-quality, full-text digitization in standard formats such as TEI, and their open-access dissemination, will alleviate. However, trying to show that based on such data, interesting results can be obtained for literary history, may be the best way to motivate continued digitization efforts. This is what this study, despite its specific thematic and methodological scope, has attempted to do. 49

## Version history and acknowledgements

The research published here was first presented at the workshop on *Computer-based Analysis of Drama* organized by Katrin Dennerlein in Munich, Germany, in March 2015 and at the *Göttingen Dialog on Digital Humanities* in April 2015. A first version of this text was submitted to *Digital Humanities Quarterly* at the end of May 2015. The author would like to thank the anonymous reviewers for their constructive comments to that version. The current version was revised in October 2016 and slightly updated in May 2017. 50

A special thank is due to Paul Fièvre who, by relentlessly building his *Théâtre classique* collection over many years, has made this research possible. 51

Work on this paper was supported by funding received from the German Federal Ministry for Research and Education (BMBF) under grant identifiers FKZ 01UG1408 and 01UG1508. 52

# Annex

All data, metadata, code and charts used in this study are available from GitHub, see: https://github.com/cligs/projects (folder 2015/gddh). 53

- Document-level metadata for the text collection
- The topic model used in this study (topics with words)
- Classification performance based on word frequencies
- Classification performance based on topic probabilities

## Notes

[1] [Schmidt 2014] has used Topic Modeling for the analysis of screenplays of TV shows, which may be considered a genre related to the more traditional dramatic texts considered here.

[2] The metadata table used can be found in the annex to this paper.

[3] The tmw toolchain is being developed at https://github.com/cligs/tmw (DOI: 10.5281/zenodo.439975). It uses lxml to read the XML-TEI encoded files, calls TreeTagger and Mallet via the subprocess module, adapts code by Allen Riddell for aggregating per-segment topic scores, relies heavily on pandas and scipy for data management and statistical analyses, and uses the word_cloud, seaborn and pygal modules for visualization.

[4] The vast majority of the text segments have a length of around 900 to 1100 word tokens, with only a small number of shorter segments originating from the end of some of the plays. This choice has been inspired by reports by [Jockers 2013, 134] that in his corpus of novels, text segments of 1000 words with arbitrary boundaries produced the best, i.e. the most interpretable topics. However, the relation between text segment length and formally measured topic coherence scores, for example using one of the measures implemented in Palmetto (see [Röder et al. 2015]) has not been assessed in this study. In an earlier iteration of this research, structurally-motivated text segments have been used by splitting the plays on the act and scene boundaries and merging very short scenes while splitting longer ones. An argument can be made for respecting the original act and scene boundaries which can be assumed to be, in the majority of cases, locations of thematic shift. However, it turns out that whether the segments have been split at arbitrary or structurally-motivated points does not affect the results as much as one might expect. Therefore, the decision has been made to use the more straightforward option here.

[5] The PRESTO language model and tagset have been developed for sixteenth- and seventeenth-century French. Eighteenth-century French, especially in tragedies, retains many of the characteristics of this earlier period, so that the tagger can be expected to work equally well with that part of the corpus. An informal sampling test revealed that for the task of identifying the nouns, verbs and adjectives, the tagging step reaches a very satisfactory level of accuracy (F1-score: 0.945), with excellent recall (extremely few false negatives) and very good precision (very few false positives). Precision is then increased in the following, rule-based lemma selection step, as it involves a short stoplist for false positives. Due to this two-step process, the list of lemmas ultimately used to build the MALLET corpus has a very high accuracy.

[6] Optimizing the parameters by optimizing some measure of model quality would have been another empirical way to solve this issue. Topic coherence measures such as the ones implemented in Palmetto (see [Röder et al. 2015]), in particular, would be interesting for this and many other purposes. The optimal parameters could then be found by maximizing both average topic coherence scores (to obtain maximally semantically coherent topics) and overall topic word vector distances (to obtain maximally distinct topics) as a function of the parameters. However, doing so independently of a specific task may be a misleading strategy in the case at hand. Indeed, it is likely that it results in a high number of very coherent topics with each of them being specific to a small number of documents, rather than serving the purpose of finding topics characteristics of large groups of plays, such as plays belonging to a given subgenre.

[7] Apart from the model selection process described here, the effect of the choice of number of topics can be described as follows: If the number of topics is set to a smaller value, several distinct and well-defined topics fail to be included among the results and several topics become aggregates of more than one semantic field. If it is set to a larger value, an overly large number of very similar topics emerge. The number of topics chosen may also interact with the length of the text segments used.

[8] See https://github.com/cligs/projects (folder "2015/gddh/"). This repository is regularly archived for long-term availability on Zenodo.org. The relevant release for this paper is v.0.3.1 (DOI: 10.5281/zenodo.439982).

[9] This has not been assessed here systematically and could be investigated in the future using hierarchical Topic Modeling or by clustering the topics based on the similarity of their word vectors.

[10] An alternative to this strategy is to perform clustering of topics with the scores of each topic in each genre as the features (not shown). This yields similar results but also shows that each genre seems to have two distinct groups of characteristic topics: those that are highly distinctive of them, and those which, although on a significantly lower level, also tend to be more associated with one genre than with the

two others.

[11] PCA has been performed using the sklearn package for Python. In our case, the first three principal components, together, account for only 22% of the variation in the original 60 topic dimensions. The cumulated proportion of the variance reaches 60% at 19 dimensions and 80% at 33 dimensions out of 60. Although PCA performs on data that is already the result of a dimensionality reduction, there are still correlations in the topic distribution data that are captured by the PCA.

[12] This observation can be confirmed mathematically by computing the correlation between genre categories and positions on the first principal component (t-test correlation is very strong at -0.81, and highly significant at p < 0,0001).

[13] Several plays which in an earlier iteration of this research appeared to have been placed in a subcluster dominated by another genre turned out to be mislabeled in the source of the data used here. This has been corrected in the meantime.